\documentclass[11pt,a4paper]{article}
\usepackage[OT1]{fontenc}
\usepackage[hyperref]{emnlp-ijcnlp-2019}
\usepackage{times}
\usepackage{latexsym}
\usepackage{multirow}

\usepackage{url}
\usepackage{graphicx}
\usepackage{algorithm}
\usepackage{amsfonts}
\usepackage{amsmath}
\usepackage{algpseudocode}
\usepackage[hang,flushmargin]{footmisc}
\usepackage{natbib}

\aclfinalcopy % Uncomment this line for the final submission

\title{Denoising based Sequence-to-Sequence Pre-training for Text Generation}

\author{Liang Wang$^1$, Wei Zhao$^1$, Ruoyu Jia$^1$, Sujian Li$^2$, Jingming Liu$^1$ \\
  $^1$Yuanfudao AI Lab, Beijing, China \\
  $^2$Key Laboratory of Computational Linguistics, Peking University, MOE, China \\
  {\tt \{wangliang01,zhaowei01,jiary,liujm\}@fenbi.com} \\
  \tt lisujian@pku.edu.cn\\}
\date{}

\begin{document}
\maketitle
\begin{abstract}
This paper presents a new sequence-to-sequence (seq2seq) pre-training method
PoDA (\textbf{P}re-training \textbf{o}f \textbf{D}enoising \textbf{A}utoencoders),
which learns representations suitable for text generation tasks.
Unlike encoder-only (e.g., BERT) or decoder-only (e.g., OpenAI GPT) pre-training approaches,
PoDA jointly pre-trains
both the encoder and decoder by denoising the noise-corrupted text,
and it also has the advantage of keeping the network architecture unchanged in the subsequent fine-tuning stage.
Meanwhile,
we design a hybrid model of Transformer
and pointer-generator networks as the backbone architecture for PoDA.
We conduct experiments on two text generation tasks:
abstractive summarization,
and grammatical error correction.
Results on four datasets show that PoDA can improve model performance over strong baselines
without using any task-specific techniques
and significantly speed up convergence.
\footnote{The code and pre-trained models are available at \url{https://github.com/yuantiku/PoDA}.}
\end{abstract}

\section{Introduction}
Methods based on unsupervised pre-training
and supervised fine-tuning for NLP
have achieved phenomenal successes in the last two years.
Most of the proposed methods in the literature
choose language modeling or its variant as the pre-training task.
After the pre-training stage,
ELMo ~\cite{peters2018deep} and CoVe ~\cite{mccann2017learned}
directly use the learned representations
as additional features for downstream tasks,
while BERT ~\cite{devlin2018bert}, ULMFiT ~\cite{howard2018universal},
XLM ~\cite{lample2019cross}, and OpenAI GPT ~\cite{radford2018improving,radford2019language}
require fine-tuning both pre-trained parameters
and task-specific parameters on labeled data.
The state-of-the-art performances have been significantly advanced
for classification and sequence labeling tasks,
such as natural language inference ~\cite{bowman2015large},
named-entity recognition,
SQuAD question answering ~\cite{rajpurkar2016squad} etc.

However,
little attention has been paid to pre-training
for seq2seq text generation ~\cite{sutskever2014sequence}.
A typical seq2seq network consists of a bidirectional encoder,
a unidirectional decoder
and attention between the encoder and decoder.
Previous work mainly focuses on encoder-only or decoder-only pre-training.
For example,
BERT pre-trains a bidirectional encoder,
and OpenAI GPT pre-trains a language model
which is essentially a unidirectional decoder.
~\newcite{ramachandran2016unsupervised} propose to train two independent language models
for the encoder and decoder respectively.
All of the aforementioned methods are only able to partially pre-train the seq2seq networks,
and therefore are unable to unleash the full potential
of transfer learning for text generation.

\begin{figure*}[ht]
\begin{center}
 \includegraphics[width=0.8\linewidth]{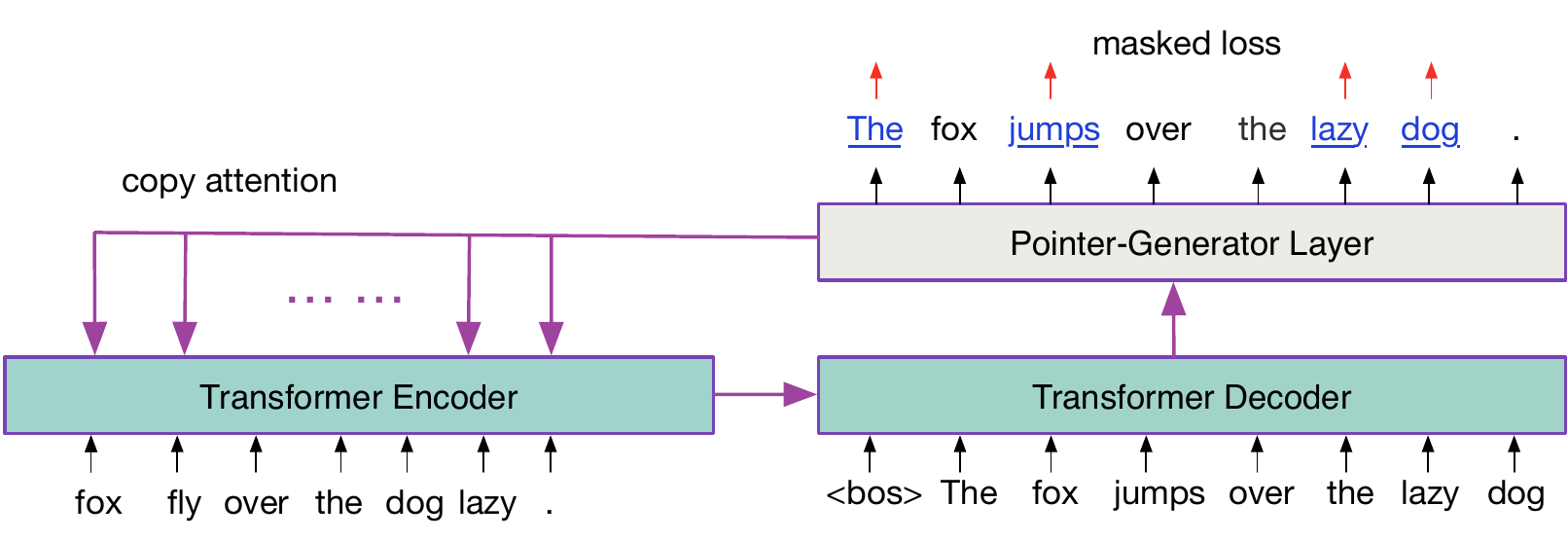}
 \caption{PoDA model architecture.
 The masked loss is calculated only for the blue underlined words.
 ``\textless bos\textgreater'' is a special begin-of-sequence padding symbol.
 The example input-output pair is explained in Section ~\ref{section:noising}.}
 \label{fig:transformer_pg}
\end{center}
\end{figure*}

In this paper,
we present PoDA,
a denoising based pre-training method that is able to jointly
pre-train all components of seq2seq networks.
Like denoising autoencoders,
PoDA works by denoising the noise-corrupted text sequences.
Any noising function that fits in the seq2seq framework can be used.
We experiment with three types of noises:
randomly shuffle, delete or replace the words in a given sequence.
It is noted PoDA is simple, easy-to-implement
and applicable to virtually all seq2seq architectures,
including ConvS2S ~\cite{gehring2017convolutional} and Transformer ~\cite{vaswani2017attention}.
Here, we adopt the hybrid architecture of Transformer
and pointer-generator networks ~\cite{see2017get}.
Transformer is effective at modeling long-distance dependencies,
highly parallelizable and demonstrates good performance empirically.
Pointer-generator network incorporates copying mechanism ~\cite{gu2016incorporating,gulcehre2016pointing}
which is helpful for most text generation tasks.

The text corpora used for pre-training are the Billion Word Benchmark ~\cite{chelba2013one}
and English Wikipedia,
both of which are publicly available
and consists of nearly $2.99$ billion words in total.
We conduct experiments on two abstractive summarization datasets
(CNN/Daily Mail ~\cite{see2017get} and Gigaword ~\cite{rush2015neural}),
and two grammatical error correction datasets
(CoNLL-2014 ~\cite{ng2014conll} and JFLEG ~\cite{napoles2017jfleg}).
With simple maximum likelihood training
and no task-specific techniques,
PoDA achieves superior or comparable performance against state-of-the-art systems
and speeds up convergence for all four datasets.

\section{Method}

\subsection{Model Architecture} \label{section:model}
First,
we design a seq2seq model as the backbone architecture of our proposed pre-training method,
which is a combination of Transformer
and pointer-generator networks,
as shown in Figure ~\ref{fig:transformer_pg}.

The input representations are the sum of word embeddings and sinusoidal positional encodings.
Both the Transformer encoder and the decoder consist of $6$ layers of transformer blocks,
and each block is a multi-head self-attention layer
followed by one layer of positionwise feedforward network.

For the output layer,
we use a pointer-generator layer to allow
both copying from the input sequence and generation from a fixed vocabulary.
The implementation is detailed in Appendix.

As a side note,
we want to point out that
the seq2seq architecture is not limited to the one we propose
and other networks such as ConvS2S, RNN-based seq2seq models are also applicable.
Pointer-generator networks are also not the only solution for handling out-of-vocabulary(OOV) words,
and subword-based methods such as sentencepiece ~\cite{kudo2018sentencepiece} can be used
at the cost of making the input and output sequences longer.

\subsection{Noising and Denoising}\label{section:noising}
Similar to denoising autoencoders,
PoDA involves two parts:
noising and denoising.
The noising part corrupts a given word sequence $\mathbf{x}=\{\mathbf{x}_i\}_{i=1}^{n}$
and gets a noisy word sequence $\mathbf{x'}=\{\mathbf{x'}_i\}_{i=1}^{n'}$.
The denoising part tries to recover $\mathbf{x}$ given $\mathbf{x'}$ using a seq2seq model.

We use three noising functions:
randomly shuffle, delete or replace the words in $\mathbf{x}$.
The details are shown in Algorithm ~\ref{algo:noising},
where $N(0, \sigma)$ is a gaussian distribution with mean $0$ and variance $\sigma$.
$B(p)$ is a Bernoulli distribution,
and $Beta(\alpha, \beta)$ is a beta distribution serving as the prior for $B(p)$.
Take function \texttt{DELETE} (line $10$ to line $15$ in Algorithm ~\ref{algo:noising}) as an example,
it first samples a Bernoulli distribution with expectation $p$ from $Beta(\alpha, \beta)$,
then each word is deleted with probability $p$.
The usage of $Beta(\alpha, \beta)$ prior can make the model robust to different degrees of noises.

We exemplify the operations above in Figure ~\ref{fig:transformer_pg}.
The original word sequence $\mathbf{x} =$\emph{``The fox jumps over the lazy dog .''},
after three noising operations:
delete \emph{``The''}, replace \emph{``jumps''} with \emph{``fly''}
and swap \emph{``lazy''} and \emph{``dog''},
we get the noisy word sequence $\mathbf{x'} =$\emph{``fox fly over the dog lazy .''}.

\begin{algorithm}[ht]
\caption{The Noising Algorithm
\newline \textbf{Input}: $\mathbf{x}$ is a sequence of words
\newline \text{\ \ \ \ \ \ \ \ \ \ } $\alpha, \beta, \sigma$ are hyperparameters}
    \begin{algorithmic}[1]
        \Function{noising}{$\mathbf{x}$} \label{algo:noising}
        \State Apply \texttt{SHUFFLE}, \texttt{DELETE}, \texttt{REPLACE} to $\mathbf{x}$ in random order
        \EndFunction
        \newline
        \Function{shuffle}{$\mathbf{x}$}
        \For{$i \gets 1\ to\ len(\mathbf{x})$}
            \State $indices[i] \gets i + \delta_i \sim N(0, \sigma)$
        \EndFor
        \State Rearrange $\mathbf{x}$ based on $argsort(indices)$
        \EndFunction
        \Function{delete}{$\mathbf{x}$}
            \State Sample $ p \sim Beta(\alpha, \beta)$
            \For{$w\ in\ \mathbf{x}$}
                \State Delete $w$ if $\mu\sim$$B(p)$ is $1$
            \EndFor
        \EndFunction
        \Function{replace}{$\mathbf{x}$}
            \State Sample $ p \sim Beta(\alpha, \beta)$
            \For{$w\ in\ \mathbf{x}$}
                \State Replace $w$ with $w'$ sampled from unigram distribution if $\mu\sim$$B(p)$ is $1$
            \EndFor
        \EndFunction
    \end{algorithmic}
\end{algorithm}

The denoising part maximizes the conditional probability $p(\mathbf{x}|\mathbf{x'})$,
which can be factorized as:
\begin{equation} \label{equation:fact}
    p(\mathbf{x}|\mathbf{x'}) = \Pi_{i=1}^{n}p(\mathbf{x}_i|\mathbf{x'}, \mathbf{x}_{<i})
\end{equation}

When predicting $\mathbf{x}_i$,
it is conditioned on the noise-corrupted full context $\mathbf{x'}$
and the clean left context $\mathbf{x}_{<i}$.
This shows that
our seq2seq formulation is capable of unifying both encoder-only pre-training
and decoder-only pre-training methods,
since a bidirectional language model used by BERT can be seen as simulating $p(\mathbf{x}_i|\mathbf{x'})$ ,
while a traditional unidirectional language model used by OpenAI GPT as resembling $p(\mathbf{x}_i|\mathbf{x}_{<i})$.

Like BERT,
we add a mask to the target sequence when computing the loss function.
To force the model to learn meaningful representations,
instead of copying from the input most of the time,
the positions where the corresponding words are corrupted in the input are kept.
We also keep a small percentage ($3\%$) of positions where the words are not corrupted,
so that the model can learn to copy from the input when appropriate.
Then, the training loss with mask is as follows ($\mathbf{\Theta}$ is model parameters):
\begin{equation}
    L = -\sum_{i=1}^{n}mask(i)\cdot\log p(\mathbf{x}_i|\mathbf{x'}, \mathbf{x}_{<i}, \mathbf{\Theta})
\end{equation}

Empirically,
we set $\sigma=0.5$ for Gaussian distribution.
$\alpha$ and $\beta$ are chosen to have
a Beta distribution with mean $0.15$ and standard deviation $0.03$.

\subsection{Pre-training Procedure}

\begin{table}[ht]
\centering
\begin{tabular}{c|c}
\hline
Corpus                     & \#words \\ \hline
English Wikipedia          & $2.22$B   \\
Billion Word Benchmark & $0.76$B   \\ \hline
Total                      & $2.99$B   \\ \hline
\end{tabular}
\caption{Text corpora used for pre-training.}
\label{table:pretrain_dataset}
\end{table}

For pre-training,
we use two text corpora:
the full dump of English Wikipedia\footnote{\url{https://dumps.wikimedia.org/}}
and the Billion Word Benchmark\footnote{\url{http://www.statmt.org/lm-benchmark/},
we do not use BooksCorpus ~\cite{zhu2015aligning} used by BERT, because it is not publicly available now. },
as shown in Table ~\ref{table:pretrain_dataset}.
For English Wikipedia,
we remove paragraphs with less than $3$ words
or more than $30\%$ OOV words,
and each paragraph is split into text segments with no more than $128$ words for each segment.
The Billion Word Benchmark is a sentence-level corpus.
Sentences with more than $500$ words are ignored during training.

The pre-training is performed on $4$ GPUs using synchronous data parallelism,
gradients are averaged across different GPUs.
Each batch on a single GPU consists of at most $3000$ tokens.
We pre-train the network for $5$ million iterations,
which is roughly $14$ epochs over the entire dataset.
The final perplexity on the validation set is about $6.8$.
Each epoch takes approximately $2$ days.
Details on the network hyperparameters and optimizers are given in Section ~\ref{section:setup}.

\subsection{Fine-tuning Procedure}
With our pre-training method,
we do not need to change the network architecture during the fine-tuning stage,
since both the pre-training task and  text generation tasks take a source sequence as input
and return a target sequence as output.
The network is initialized with pre-trained parameter values.
For fine-tuning,
the preprocessing is dataset-specific,
but the learning rate scheduling,
dropout, early stopping,
and gradient clipping are exactly the same as pre-training.

The objective function for fine-tuning is the word-level negative log-likelihood.
Here we do not use reinforcement learning to tune towards
the automatic evaluation metrics such as ROUGE ~\cite{lin2004rouge} or BLEU ~\cite{papineni2002bleu},
because it may overfit evaluation metrics
and barely show improvements in human evaluations ~\cite{wu2016google}.

\section{Experiments}

\subsection{Setup} \label{section:setup}
The network architecture used by our experiments
has $97$ million parameters.
It consists of $6$ layers of encoder blocks,
$6$ layers of decoder blocks,
and $1$ pointer-generator layer.
The hidden size of each positionwise feedforward layer is $4096$.
We use $8$ heads for all multi-head attention layers.
The vocabulary consists of the top $50k$ most frequent words (case sensitive),
and the dimension of word embedding is $512$.
We tie the parameters of encoder word embeddings, decoder word embeddings,
and the output softmax layer.
NAG (Nesterov Accelerated Gradient) optimizer is used with initial learning rate $2 \times 10^{-3}$.
Dropout of $0.2$ is applied for all self-attention layers,
positionwise feedforward layers and input embedding layers.
The gradient norm is clipped to have a maximum value of $2$.
We follow the Transformer implementation from
\emph{fairseq}\footnote{\url{https://github.com/pytorch/fairseq}}.

For task-specific fine-tuning,
unless explicitly specified,
we reuse the hyperparameters from the pre-training stage.
After each training epoch,
we compute the validation loss and halve the learning rate
whenever the validation loss stops decreasing.
The training procedure terminates if the learning rate drops below $10^{-4}$.
Exponential moving average (EMA) with decay rate $0.9995$ is used to make the training stabilized.
At inference time,
we use standard beam search decoding based on the length-normalized log-likelihood.
For ensemble models,
we use different random seeds and pre-trained checkpoints for fine-tuning.
Ensemble decoding is used by averaging the output probabilities
from different models at every decoding step.

When reporting experimental results,
\emph{``PoDA w/o pre-training''} refers to the proposed architecture in Section ~\ref{section:model}
trained only on the supervised data,
and \emph{``PoDA w/o fine-tuning''} only pre-trains on unlabeled data.
\emph{PoDA} first pre-trains a denoising autoencoder
and then fine-tunes on the supervised data.

\subsection{Abstractive Summarization}
\noindent
\textbf{Datasets }
We use two summarization datasets:
CNN/Daily Mail\footnote{We use the non-anonymized version,
which is considered to be more realistic.} ~\cite{see2017get}
and Gigaword ~\cite{rush2015neural} dataset.
The official split for training, validation, and test
is shown in Table ~\ref{table:dataset_summ}.

\begin{table}[ht]
    \centering
    \scalebox{0.95}{\begin{tabular}{c|ccc}
    \hline
    \multirow{2}{*}{\small{Corpus}} & \multicolumn{3}{c}{\small{\# of examples}} \\ \cline{2-4}
                            & \small{train}      & \small{valid}      & \small{test}     \\ \hline
    \small{CNN/Daily Mail}   &    $287,113$   &  $13,368$   &   $11,490$  \\
    \small{Gigaword}        & $3,803,957$ & $189,651$  & $1,951$ \\ \hline
    \end{tabular}}
    \caption{Dataset statistics for abstractive summarization.}
    \label{table:dataset_summ}
\end{table}

The CNN/Daily Mail dataset contains approximately $300k$ news articles
with an average of $781$ words for each article,
and each article is paired with summaries with $56$ words on average.
We use the preprocessing script\footnote{\url{https://github.com/abisee/cnn-dailymail}}
provided by ~\newcite{see2017get}.
The articles are truncated to $800$ words for both training and testing.
The summaries are truncated to $130$ words for training.

The Gigaword is a headline-generation dataset
consisting of nearly $4$ million examples.
Headline generation can be seen as a sentence summarization task.
Each example in Gigaword consists of one sentence with an average length of $31.3$ words,
which is much shorter than CNN/Daily Mail,
and one short headline with an average length of $8.3$ words.
The Gigaword dataset provided by ~\newcite{rush2015neural}
is already tokenized and lower-cased.
Since our vocabulary is case-sensitive,
such inconsistency is expected to hurt our system's performance.

\noindent
\textbf{Evaluation }
We report evaluation results in terms of of ROUGE-1, ROUGE-2 and ROUGE-L ~\cite{lin2004rouge}
using the \emph{pyrouge}\footnote{\url{https://github.com/andersjo/pyrouge}} package.
For the CNN/Daily Mail dataset,
PGNet ~\cite{see2017get},
Lead3 ~\cite{see2017get}, rnn-ext + RL ~\cite{chen2018fast},
NeuSum ~\cite{zhou2018neural} are used as baselines.
For the Gigaword dataset,
ABS+ ~\cite{rush2015neural},
CGU ~\cite{lin2018global}, FTSum ~\cite{cao2018faithful},
and R$e^3$Sum ~\cite{cao2018retrieve} are used as baselines.

\begin{table}[ht]
\centering
\scalebox{0.92}{\begin{tabular}{c|ccc}
\hline
\multirow{2}{*}{System} & \multicolumn{3}{c}{ROUGE} \\ \cline{2-4}
                        & 1       & 2      & L      \\ \hline
Lead3  & $40.34$  & $17.70$   & $36.57$      \\
PGNet &   $36.44$  & $15.66$   &   $33.42$  \\
rnn-ext + RL  &  $41.47$ &   $18.72$  &  $37.76$ \\
NeuSum  &  $41.59$ &  $19.01$ &  $37.98$  \\ \hline
\textbf{PoDA w/o pre-training}    &   $40.82$  &  $18.46$  &  $37.61$   \\
\textbf{PoDA} &  $\mathbf{41.87}$  &  $\mathbf{19.27}$  &  $\mathbf{38.54}$  \\ \hline
\end{tabular}}
\caption{ROUGE scores for CNN/Daily Mail dataset.}
\label{table:cnndm}
\end{table}

\noindent
\textbf{Results for CNN/Daily Mail }
Considering the characteristics of news articles,
baselines such as Lead3 (simply choose the first 3 sentences)
can achieve strong performance in terms of ROUGE scores,
as shown in Table ~\ref{table:cnndm}.
``rnn-ext+RL''  combines both extractive and abstractive methods
and achieves performance improvements ~\cite{chen2018fast}.
PoDA is a purely abstractive summarization system and performs stably better than all the methods.
``PoDA w/o pre-training'' only has moderate success with ROUGE-1 $40.82$, ROUGE-2 $18.46$ and ROUGE-L $37.61$.
When combined with pre-training,
PoDA establishes new state-of-the-art on CNN/Daily Mail dataset.

\begin{table}[ht]
\centering
\scalebox{0.9}{\begin{tabular}{c|ccc}
\hline
\multirow{2}{*}{System} & \multicolumn{3}{c}{ROUGE} \\ \cline{2-4}
                        & 1       & 2      & L      \\ \hline
ABS+ & $29.76$ & $11.88$ & $26.96$ \\
CGU  & $36.3$ & $18.0$ & $33.8$ \\
FTSum & $37.27$  &  $17.65$ & $34.24$  \\
R$e^3$Sum  & $37.04$ & $19.03$  & $34.46$ \\ \hline
\textbf{PoDA w/o pre-training} &   $37.24$  &  $18.28$ &   $34.53$  \\
\textbf{PoDA} &   $\mathbf{38.29}$   &  $\mathbf{19.06}$ &   $\mathbf{35.45}$ \\ \hline
\end{tabular}}
\caption{ROUGE scores for Gigaword dataset.}
\label{table:gigaword}
\end{table}

\noindent
\textbf{Results for Gigaword }
The Gigaword dataset is much larger than CNN/Daily Mail,
and this enables ``PoDA w/o pre-training'' to have competitive performance even without pre-training.
As in Table ~\ref{table:gigaword},
PoDA substantially improves ROUGE-1 from $37.24$ to $38.29 (+1.05)$,
ROUGE-2 from $18.28$ to $19.06 (+0.78)$, and ROUGE-L from $34.53$ to $35.45 (+0.92)$,
with pre-training added.
We can see that PoDA performs favorably over the state-of-the-art R$e^3$Sum method,
which utilizes unlabeled text corpora
by first retrieving and then rewriting relevant snippets.

\subsection{Grammatical Error Correction (GEC)}
\noindent
\textbf{Datasets }
GEC can also be seen as a text generation task,
where the input sequence is a sentence possibly containing some grammatical errors,
and the output is a clean and grammatical sentence.
We experiment PoDA on two GEC datasets: CoNLL-2014 ~\cite{ng2014conll} and JFLEG ~\cite{napoles2017jfleg}.
We use three public datasets for training:
Lang-8 NAIST ~\cite{mizumoto2011mining},
NUCLE ~\cite{dahlmeier2013building} and CLC FCE ~\cite{felice2014grammatical}.
The test set of CoNLL-2013 shared task is used as validation set
for the CoNLL-2014 task.
JFLEG has its own validation set.
For preprocessing,
we use NLTK\footnote{\url{https://www.nltk.org/}} to tokenize sentences,
and remove all sentence pairs without any edits in Lang-8 NAIST.
Simple spelling errors are corrected based on edit distance.
The dataset statistics are shown in Table ~\ref{table:gec_dataset}.

\begin{table}[ht]
\centering
\begin{tabular}{c|cc}
\hline
Corpus       & \#Sent Pairs & Split      \\ \hline
Lang-8 NAIST & $1,097,274$   & train      \\
NUCLE        & $57,113$      & train      \\
CLC FCE      & $32,073$      & train      \\ \hline
CoNLL-2013 test set   & $1,381$        & valid \\
JFLEG valid set  &   $754$    & valid  \\ \hline
CoNLL-2014 test set   & $1,312$        & test       \\
JFLEG test set        & $747$         & test       \\ \hline
\end{tabular}
\caption{Dataset statistics for grammatical error correction.
Due to the lack of standard preprocessing script,
the number of sentence pairs in the training set
are slightly different from previous work.}
\label{table:gec_dataset}
\end{table}

\noindent
\textbf{Evaluation }
To compare with previous work,
we use the official evaluation metrics:
MaxMatch ($M^2$) $F_{0.5}$ ~\cite{dahlmeier2012better} for CoNLL-2014
and GLEU ~\cite{napoles2015ground} for JFLEG dataset.
Both metrics are shown to correlate well with human evaluation scores.
MLConv ~\cite{chollampatt2018multilayer}, char-seq2seq ~\cite{xie2016neural},
dual-boost ~\cite{ge2018fluency},
Hybrid SMT-NMT ~\cite{grundkiewicz2018near},
NQE ~\cite{chollampatt2018neural}, and NRL ~\cite{sakaguchi2017grammatical} are used as baselines.

\begin{table}[ht]
\centering
\scalebox{0.9}{\begin{tabular}{c|ccc}
\hline
System (single) & $P$ & $R$ & $F_{0.5}$ \\ \hline
char-seq2seq  & $49.24$  & $23.77$  &  $40.56$ \\
MLConv & $60.90$ & $23.74$  & $46.38$  \\
dual-boost  &  $62.70$  & $27.69$  &  $50.04$ \\
\textbf{PoDA w/o fine-tuning} & $19.22$ & $31.73$  & $20.86$  \\
\textbf{PoDA w/o pre-training} & $65.63$ & $31.62$  & $54.01$  \\
\textbf{PoDA} & $\mathbf{70.10}$ & $\mathbf{36.88}$  & $\mathbf{59.40}$  \\ \hline
Ensemble & &  & \\ \hline
MLConv(+rerank) & $65.49$  & $33.14$  &  $54.79$ \\
SMT-NMT(+rerank) & $66.77$ & $34.49$ & $56.25$ \\
NQE  &  - & -  &  $56.52$    \\
\textbf{PoDA}  &  $\mathbf{71.01}$ & $\mathbf{37.68}$  & $\mathbf{60.34}$  \\ \hline
\end{tabular}}
\caption{Precision ($P$), recall ($R$) and $F_{0.5}$ scores for CoNLL-2014 test set.
We only list systems trained on public data.
~\newcite{ge2018reaching} reported better performance with additional 4 million
non-public sentence pairs.}
\label{table:conll2014}
\end{table}

\begin{table}[ht]
\centering
\begin{tabular}{c|ccc}
\hline
System (single) & valid & test \\ \hline
MLConv & $47.71$ &  $51.34$ \\
NRL   & $49.82$  & $53.98$  \\
dual-boost  & $51.35$  &  $56.33$  \\
\textbf{PoDA w/o fine-tuning} & $34.43$ & $36.83$ \\
\textbf{PoDA w/o pre-training}  &  $51.57$ &   $56.52$   \\
\textbf{PoDA}    &  $\mathbf{53.16}$ &  $\mathbf{59.02}$    \\ \hline
Ensemble   &   &      \\ \hline
MLConv(+rerank) & $52.48$ & $57.47$ \\
SMT-NMT(+rerank)  & -  &   $\mathbf{61.50}$   \\
\textbf{PoDA}    &  $\mathbf{53.29}$ &  $59.48$    \\ \hline
Human   & -  &  $62.38$ \\ \hline
\end{tabular}
\caption{GLEU scores for JFLEG dataset.}
\label{table:jfleg}
\end{table}

\noindent
\textbf{Results }
As shown in Table ~\ref{table:conll2014} and Table ~\ref{table:jfleg},
when trained only on the supervised data,
``PoDA w/o pre-training'' can still achieve an impressive performance
with $F_{0.5}$ score $54.01$ on CoNLL-2014 test set
and GLEU score $56.52$ on JFLEG test set,
surpassing previous state-of-the-art single model results.
This once again shows the effectiveness of the Transformer architecture.
For GEC task,
most words in the output sequence also appear in the input sequence,
pointer-generator makes it easier to learn such prior.
With denoising based pre-training,
PoDA greatly boosts the $F_{0.5}$ score
from $54.01$ to $59.40 (+5.39)$ for CoNLL-2014 dataset,
and GLEU score from $56.52$ to $59.02 (+2.50)$ for JFLEG.
By ensembling $4$ models initialized with different pre-trained checkpoints
and trained with different random seeds,
the performance can be further boosted on both datasets
($+0.94$ for CoNLL-2014 and $+0.46$ for JFLEG),
outperforming the other ensemble models such as ``Hybrid SMT-NMT''.

We also report the performance of ``PoDA w/o fine-tuning'' which does not conduct fine-tuning.
The $F_{0.5}$ score only reaches $20.86$ on CoNLL-2014 dataset
and the GLEU score is $36.83$ on JFLEG.
These results are even worse than the weakest baselines
in Table ~\ref{table:conll2014} and Table ~\ref{table:jfleg}.
The denoising based pre-training and the GEC task share some similarities
in the sense that both of them attempt to convert noisy texts to clean and grammatical texts.
However,
the poor results of ``PoDA w/o fine-tuning'' show
that PoDA cannot be seen as a simple data augmentation method for GEC.
Instead,
PoDA learns generic text representations
and requires task-specific fine-tuning.

Techniques from previous work for GEC such as language model based rerank ~\cite{chollampatt2018multilayer},
data augmentation ~\cite{ge2018fluency},
and domain adaptation ~\cite{junczys2018approaching}
can be easily incorporated.
A parallel work ~\cite{zhao2019improving} observes similar gain by
combining simpler pre-training strategy and various GEC-specific techniques.

\section{Analysis}

In the following analysis,
we only choose one task (summarization or GEC) to analyze each aspect due to space limitation.
Similar conclusions also hold for the other task.

\begin{table*}[ht]
\centering
\scalebox{1.0}{\begin{tabular}{c|l}
\hline
Source                & \begin{tabular}[c]{@{}l@{}} ivory coast striker boubacar sanogo is set to leave werder bremen \\ for french first division side saint-etienne. \end{tabular} \\ \hline
Target                & \begin{tabular}[c]{@{}l@{}} sanogo set to sign for saint-etienne \end{tabular} \\ \hline
PoDA w/o pre-training & \begin{tabular}[c]{@{}l@{}} ivory coast striker sanogo set to join saint-etienne \end{tabular} \\ \hline
PoDA                  & \begin{tabular}[c]{@{}l@{}} sanogo set to leave bremen for saint-etienne  \end{tabular}   \\ \hline \hline
Source                & \begin{tabular}[c]{@{}l@{}} thailand's battered economy should start to bottom out in the first quarter \\ of \#\#\#\# provided the government's efforts are n't neutralized \\ by outside developments, a news report said monday. \end{tabular} \\ \hline
Target                & \begin{tabular}[c]{@{}l@{}} economic crisis to bottom out early next year minister says \end{tabular}   \\ \hline
PoDA w/o pre-training & \begin{tabular}[c]{@{}l@{}} thai economy expected to start to bottom out in UNK \end{tabular} \\ \hline
PoDA                  & \begin{tabular}[c]{@{}l@{}} thai economy to start to bottom out in first quarter \end{tabular} \\ \hline
\end{tabular}}
\caption{Examples of generated summaries from Gigaword dataset.}
\label{table:giga_examples}
\end{table*}

\subsection{Linguistic Quality Analysis}

In Table ~\ref{table:giga_examples},
we show some generated summaries by PoDA from Gigaword dataset.
In the first example,
PoDA successfully deletes the relatively unimportant modifier ``ivory coast striker'',
and keeps the picture complete by including both ``bremen'' and ``saint-etienne'' in the summary.
In the second example,
``PoDA w/o pre-training'' misses an important date (``first quarter'') of the event ``economic crisis''.
Both examples show our model is able to identify the important text snippets in the source sequence
and organize them into a short and fluent summary.

More example outputs by PoDA are listed in Appendix.

\subsection{Effects of the Number of Pre-training Iterations}

\begin{figure}[ht]
\begin{center}
 \includegraphics[width=0.9\linewidth]{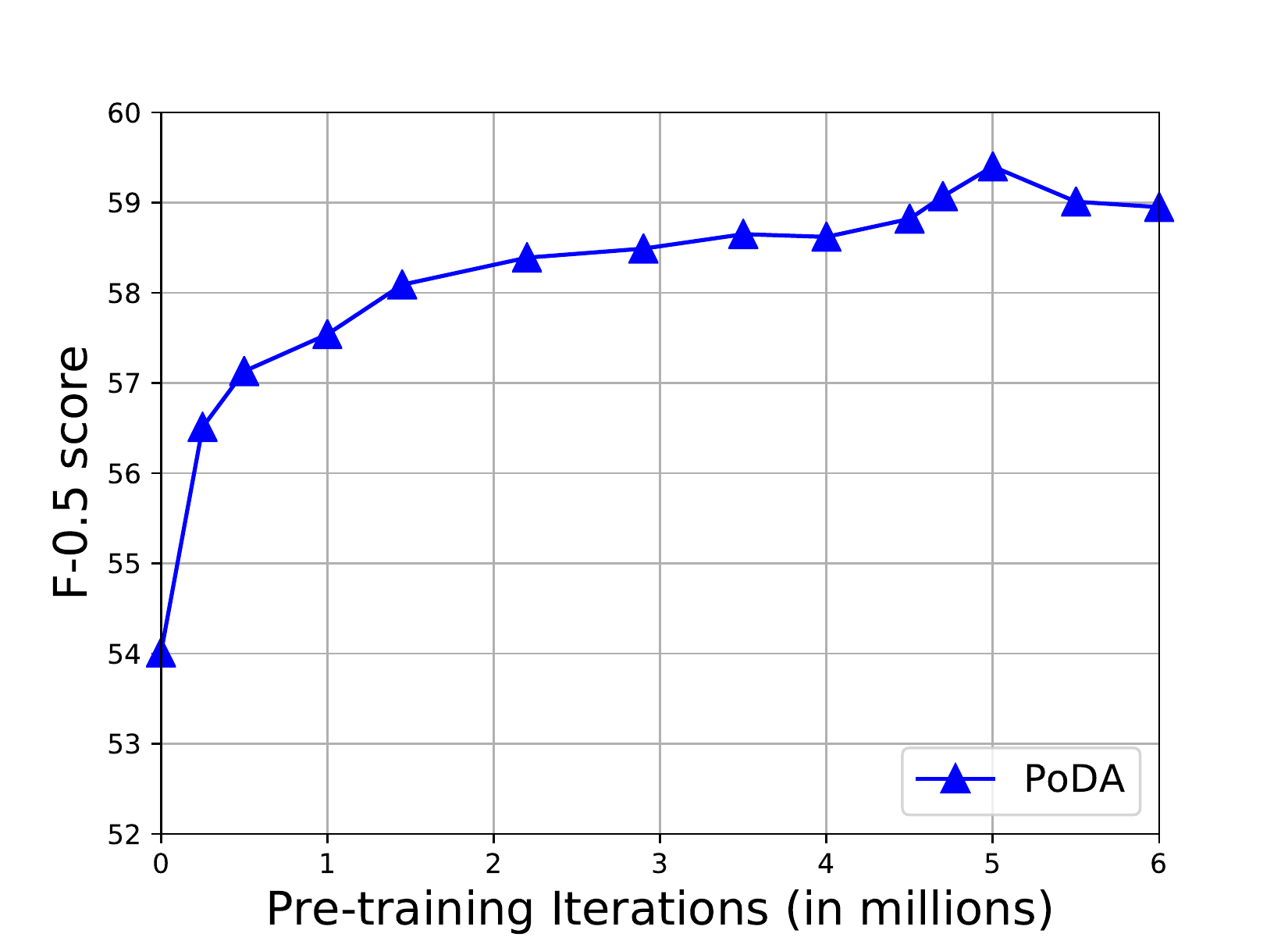}
 \caption{$F_{0.5}$ score on CoNLL-2014 test set with respect to the number of pre-training iterations.}
 \label{fig:model}
\end{center}
\end{figure}

In Figure ~\ref{fig:model},
we show the $F_{0.5}$ score on CoNLL-2014 dataset
when the model is initialized with different pre-trained checkpoints.
Though the $F_{0.5}$ score has some fluctuations due to the random factors
in training neural networks and the limited size of the test set,
the overall trend is very clear:
the $F_{0.5}$ score first improves greatly
and then keeps a slow improvement after about $1$ million iterations,
from $54$ at the very beginning to $59$ after convergence.

\subsection{Effects of Pre-trained Encoder and Decoder}

To show the effectiveness of the pre-trained encoder and decoder,
we train the model by only using the encoder-side pre-trained parameters
(``w/o pre-trained decoder'')
or decoder-side pre-trained parameters (``w/o pre-trained encoder'')
We do not compare with pre-trained encoder from BERT or pre-trained decoder from OpenAI GPT,
mainly because the corresponding model capacity, tokenization
and text corpora used for pre-training are very different.

\begin{table}[ht]
\centering
\scalebox{0.92}{\begin{tabular}{l|ccc}
\hline
\multicolumn{1}{c|}{System}    & $P$ & $R$ & $F_{0.5}$ \\ \hline
\multicolumn{1}{c|}{Fully pre-trained} & $\mathbf{70.10}$ & $\mathbf{36.88}$  & $\mathbf{59.40}$ \\ \hline
w/o pre-trained encoder      & $66.14$  & $34.67$  &  $55.98$  \\
w/o pre-trained decoder      & $66.62$  & $36.10$  &  $56.98$    \\
w/o pre-training             & $65.63$ & $31.62$  & $54.01$ \\ \hline
\end{tabular}}
\caption{Ablations for pre-trained encoder and decoder on CoNLL-2014 test set.}
\label{table:ablation}
\end{table}

Table ~\ref{table:ablation} shows that the performance degrades by a large margin
if the network is only partially pre-trained.
The pre-trained encoder ($-3.42$ drop in $F_{0.5}$) is more important
than the pre-trained decoder ($-2.42$ drop in $F_{0.5}$).

\subsection{Effects of Dataset Size}

We also conduct ablations in few-shot learning settings
to see how the performance changes
when the model only accesses a small percentage of labeled data.
We randomly sample $10^3$ to $10^5$ training examples from the Gigaword dataset
and train ``PoDA w/o pre-training'' and PoDA (with pre-training)
using exactly the same hyperparameters.

\begin{figure}[ht]
\begin{center}
 \includegraphics[width=0.95\linewidth]{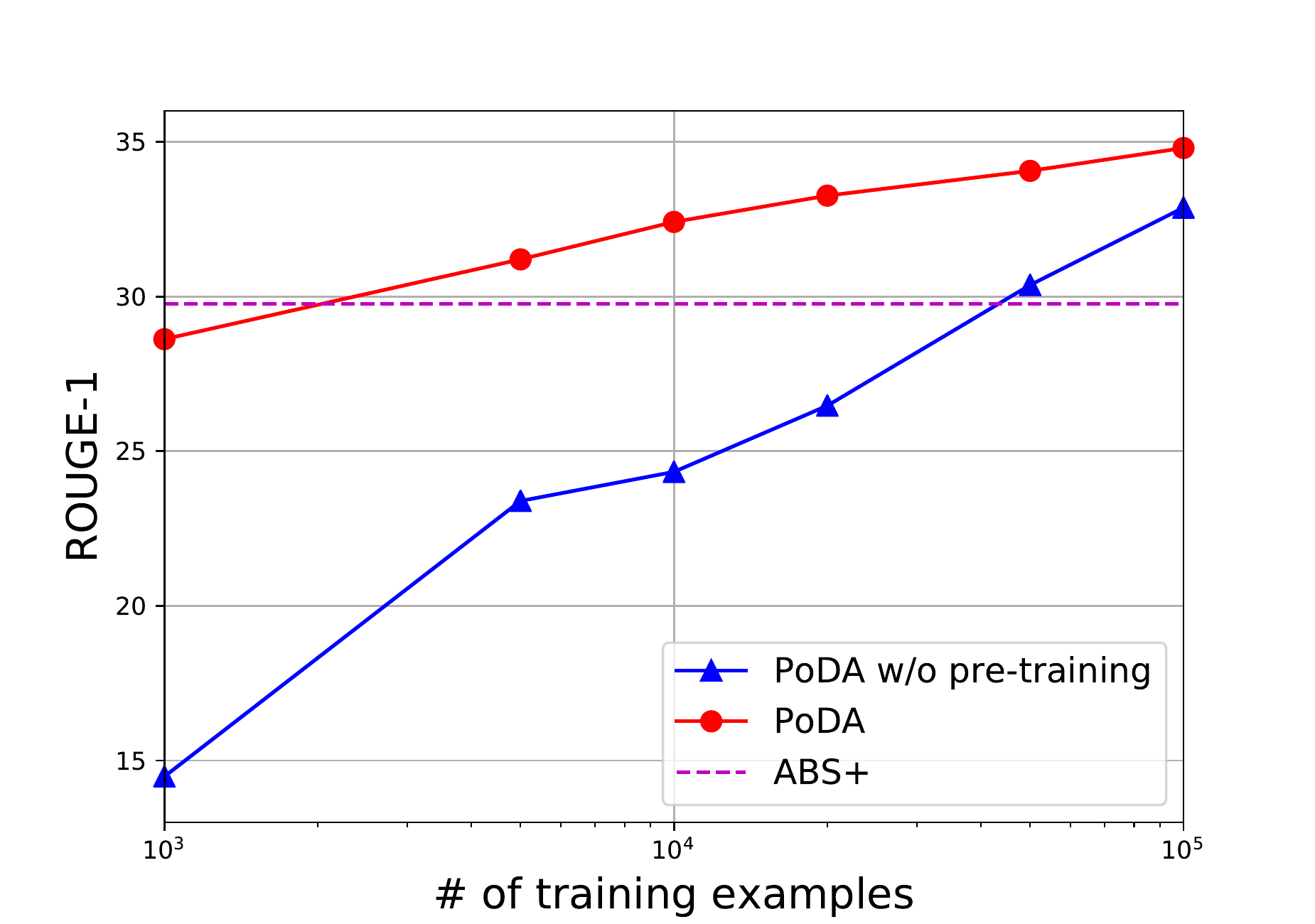}
 \caption{ROUGE-1 on Gigaword test set with respect to the number of training examples.
 ABS+ is a baseline method from ~\newcite{rush2015neural} using attention.
 The x-axis is in log scale.}
 \label{fig:dataset_size}
\end{center}
\end{figure}

\begin{figure*}[ht]
\begin{center}
 \includegraphics[width=0.95\linewidth]{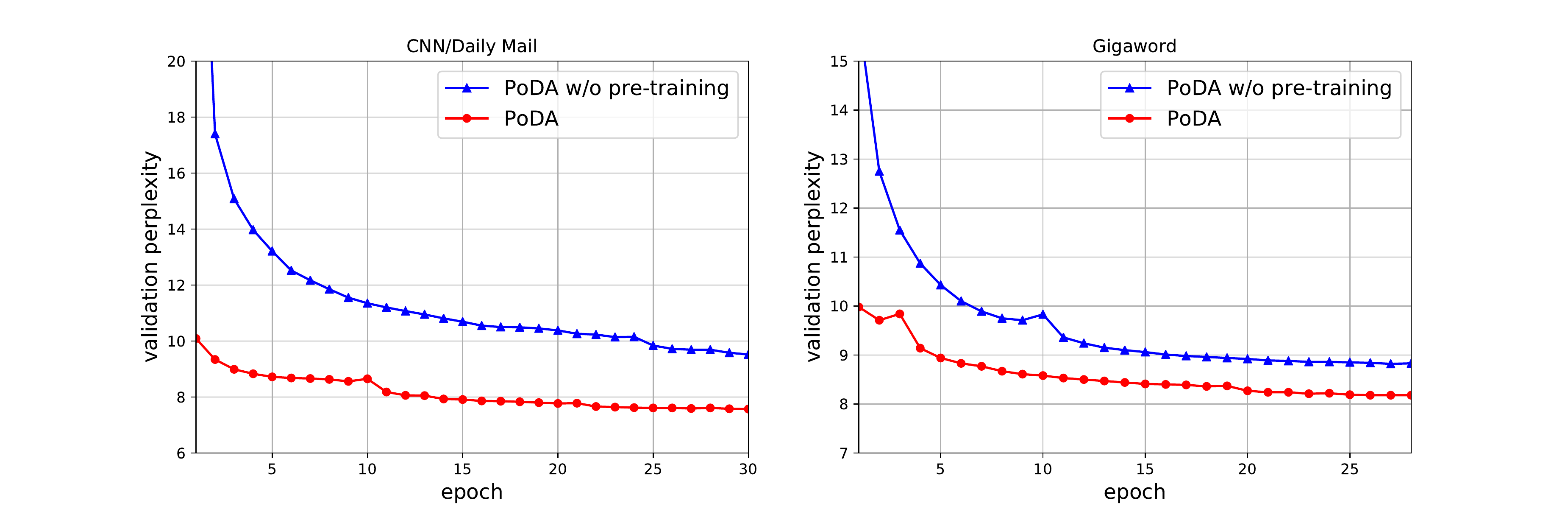}
 \caption{Validation perplexity with respect to the training epochs on CNN/Daily Mail and Gigaword datasets.
 Perplexity is related to the vocabulary size,
 so the values here are not comparable with previous work.}
 \label{fig:epoch}
\end{center}
\end{figure*}

Figure \ref{fig:dataset_size} shows the ROUGE-1 score on Gigaword test set.
With only $10^3$ training examples,
PoDA reaches a reasonably good performance
comparable to ABS+ (an attention-based system trained on nearly 4 million examples).
With more labeled data available,
the performance gap between ``PoDA w/o pre-training'' and PoDA
slowly decreases from $15$ to $2$ in Figure ~\ref{fig:dataset_size}.
However,
the pre-training still helps even
when the models are trained on the full dataset (shown in Table ~\ref{table:gigaword}).

\subsection{Convergence Analysis}
Pre-training can not only achieve better final performance
but also helps the model converge faster.
In Figure ~\ref{fig:epoch},
we show the validation perplexity after each training epoch
for both ``PoDA w/o pre-training'' and PoDA.

We can clearly see that
the validation perplexity of PoDA
is consistently lower than that of ``PoDA w/o pre-training'',
especially at the first few epochs.
After 5 epochs,
PoDA can arrive at a validation perplexity
that ``PoDA w/o pre-training'' usually takes 30 or more epochs
for both CNN/Daily Mail and Gigaword datasets.
This nice property is particularly helpful
when the computational resources are limited.
Other pre-training methods such as BERT also demonstrate similar behaviors.

\section{Related Work}
\textbf{Network Pre-training }
The idea of pre-training neural networks dates back to
the early days of deep learning.
~\newcite{bengio2007greedy} proposed layer-wise pre-training for
deep belief networks (DBN) to tackle the difficulty
of training deep neural networks based on a reconstruction objective.
~\cite{erhan2010does,dahl2012context} showed
the effectiveness of pre-training for tasks such as speech recognition.
In the area of computer vision,
using ImageNet pre-trained models have become a standard practice.
In NLP community,
using pre-trained word embeddings
is the most popular way to transfer knowledge from the unlabeled corpus.
There are also work on semi-supervised sequence learning ~\cite{dai2015semi,peters2017semi}
attempting to incorporate language modeling as an auxiliary task.
Recently,
several pre-training methods based on language models are presented,
such as ELMo ~\cite{peters2018deep}, OpenAI GPT ~\cite{radford2018improving},
BERT ~\cite{devlin2018bert}, XLM ~\cite{lample2019cross} etc.
The combination of more compute, larger model capacity
and large-scale text corpora
lead to significant improvements
on NLP benchmarks ~\cite{wang2018glue}.
\newline

\noindent
\textbf{Autoencoders }
have long been used for representation learning of images ~\cite{vincent2010stacked}
and text ~\cite{li2015hierarchical}.
However,
precisely reconstructing the clean input is probably too easy for high-capacity models.
Sparse autoencoders ~\cite{deng2013sparse}, contractive autoencoders ~\cite{rifai2011contractive},
and denoising autoencoders ~\cite{vincent2010stacked}
are several popular variants.
Denoising autoencoders (DA) are shown to be able to
learn better representations for downstream tasks
~\cite{vincent2010stacked,vincent2008extracting,hill2016learning}.
~\newcite{freitag2018unsupervised} use seq2seq DAs
for unsupervised natural language generation in dialogue,
and ~\cite{kim2018improving} propose to improve the quality of machine translation with DAs.
\newline

\noindent
\textbf{Text Generation }
covers a wide spectrum of NLP tasks,
including machine translation ~\cite{wu2016google},
summarization ~\cite{see2017get}, response generation ~\cite{vinyals2015neural},
paraphrase generation, grammatical error correction etc.
Early studies on text generation mainly adopt template-based ~\cite{reiter2000building}
or example-based ~\cite{watanabe1998pattern} methods.
With the emergence of deep learning for NLP,
seq2seq models ~\cite{sutskever2014sequence} become a popular choice for text generation tasks
and show better performance in terms of both automatic evaluation metrics
and human evaluations ~\cite{wu2016google}.
There are also studies focusing on text generation from structured data
such as SQL-to-text ~\cite{xu2018sql}.
Previous pre-training for text generation is usually done by
independently pre-training encoder-side or decoder-side language models ~\cite{ramachandran2016unsupervised}.
Concurrent to our work,
~\citeauthor{edunov2019pre} augment encoder representation with ELMo-style models,
MASS ~\cite{song2019mass} masks continuous text fragments for pre-training,
and UNILM ~\cite{dong2019unified} proposes to pre-train for both language understanding and generation tasks.

\section{Conclusion}
This paper presents a new transfer learning approach for seq2seq text generation named PoDA.
It involves two steps:
first,
pre-train a customized seq2seq denoising autoencoder on large-scale unlabeled text corpora;
then,
fine-tune on in-domain labeled data.
The pre-training step is independent of downstream tasks
and jointly learns both encoder and decoder representations.
PoDA is simple, intuitive
and doesn't require changing network architecture during the fine-tuning stage.
Experiments on several abstractive summarization and grammatical error correction datasets demonstrate
that PoDA leads to better performance and faster convergence.

For future work,
we would like to validate our model on other tasks such as response generation,
explore more effective unsupervised sequence-to-sequence pre-training methods,
and handle cross-lingual tasks such as machine translation.

\section*{Acknowledgments}
We want to thank three anonymous reviewers for their valuable comments,
and EMNLP-IJCNLP 2019 organizers for their efforts.

\bibliography{emnlp-ijcnlp-2019}
\bibliographystyle{acl_natbib}

\appendix

\section{Supplemental Material}

\subsection{Implementation of Pointer-Generator Layer}

The pointer-generator layer calculates a probability distribution
over the union of a fixed vocabulary $\mathbf{V}$ and the words in the input sequence $\{\mathbf{x}_i\}_{i=1}^n$.

For a word $w$,
the probability $p(w)$ can be calculated as follows:

\begin{equation}
\begin{aligned}
\boldsymbol{\alpha} & = MultiHeadAttention(\mathbf{h}_{dec}^t, \mathbf{h}_{enc}) \\
p_{gen} & = \sigma(\mathbf{W}_1 AttPooling(\mathbf{h}_{enc}, \boldsymbol{\alpha})) \\
p(w) & = p_{gen} p_{v}(w) + (1 - p_{gen})\sum_{i:\mathbf{x'}_i=w}\boldsymbol{\alpha}_i
\end{aligned}
\end{equation}

\begin{table*}[ht]
\centering
\scalebox{1.0}{\begin{tabular}{c|l}
\hline
Source                & \begin{tabular}[c]{@{}l@{}} I do think there is difference in it and I believe many of us will agree. \end{tabular} \\ \hline
Target                & \begin{tabular}[c]{@{}l@{}} I do think there is \textbf{a difference} and I believe many of us will agree. \end{tabular} \\ \hline
PoDA w/o pre-training & \begin{tabular}[c]{@{}l@{}} I do think there is \textbf{difference in it} and I believe many of us will agree. \end{tabular} \\ \hline
PoDA                  & \begin{tabular}[c]{@{}l@{}} I do think there is \textbf{a difference in it} and I believe many of us will agree. \end{tabular}   \\ \hline \hline
Source                & \begin{tabular}[c]{@{}l@{}} Almost all students and young adults possess the Facebook or Twitter account. \end{tabular} \\ \hline

Target                & \begin{tabular}[c]{@{}l@{}} Almost all students and young adults possess \textbf{a Facebook or Twitter account}. \end{tabular}   \\ \hline
PoDA w/o pre-training & \begin{tabular}[c]{@{}l@{}} Almost all students and young adults possess \textbf{Facebook or Twitter accounts}. \end{tabular} \\ \hline
PoDA                  & \begin{tabular}[c]{@{}l@{}} Almost all students and young adults possess \textbf{a Facebook or Twitter account}. \end{tabular} \\ \hline
\end{tabular}}
\caption{Examples of corrected sentences from CoNLL-2014 dataset.
The important text snippets are highlighted with bold font.}
\label{table:conll_examples}
\end{table*}

\begin{table*}[ht]
\centering
\scalebox{0.95}{\begin{tabular}{c|l}
\hline
Source                & \begin{tabular}[c]{@{}l@{}} But, on the contrary, he argues that fluoride also some disadvantage.  \end{tabular} \\ \hline
Target                & \begin{tabular}[c]{@{}l@{}} But, on the contrary, he argues that fluoride also \textbf{has some disadvantages}. \end{tabular} \\ \hline
PoDA w/o pre-training & \begin{tabular}[c]{@{}l@{}} But, on the contrary, he argues that \textbf{there are also some disadvantages}. \end{tabular} \\ \hline
PoDA                  & \begin{tabular}[c]{@{}l@{}} But, on the contrary, he argues that fluoride also \textbf{has some disadvantages}.  \end{tabular}   \\ \hline \hline
Source                & \begin{tabular}[c]{@{}l@{}} Such people impressed other people through their strong well and divoution to duty.\end{tabular} \\ \hline
Target                & \begin{tabular}[c]{@{}l@{}} Such people impressed others through their \textbf{strong will and devotion to duty}. \end{tabular}   \\ \hline
PoDA w/o pre-training & \begin{tabular}[c]{@{}l@{}} Such people impressed other people through their \textbf{strong and divoution to duty}.\end{tabular} \\ \hline
PoDA                  & \begin{tabular}[c]{@{}l@{}} Such people impressed other people through their \textbf{strong will and devotion to duty}.\end{tabular} \\ \hline
\end{tabular}}
\caption{Examples of corrected sentences from JFLEG dataset.
The important text snippets are highlighted with bold font.}
\label{table:jfleg_examples}
\end{table*}

\begin{table*}[ht]
\centering
\scalebox{1.0}{\begin{tabular}{c|l}
\hline
Input                             & \multicolumn{1}{c}{Output samples}                                                                                                                                                                                                                                                     \\ \hline
the best university in the world. & \begin{tabular}[c]{@{}l@{}}Harvard is considered the 10th best university in the world.\\ I believe that's the best university in the world.\\ Nevada offers the best continuing education in the whole world.\\ Greek is becoming the best university in the world.\end{tabular} \\ \hline
The meaning of life is. & \begin{tabular}[c]{@{}l@{}}The meaning of daily life is unclear.\\ ``The real meaning of daily life is peaceful.''\\ The underlying meaning of your life is lost forever.\\ The immediate meaning of our political life is undeniable.\end{tabular}  \\ \hline
\end{tabular}}
\caption{Some input-output samples by our pre-trained PoDA models.
We input some incomplete sentences and output the generated sentences
by sampling the output distribution
of pre-trained PoDA models at each timestep.
We can see that PoDA successfully transforms the inputs to coherent and grammatical sentences,
though the statements entailed by these output sentences are not always correct.}
\label{table:samples}
\end{table*}

$\mathbf{h}_{enc}$ is the top layer output of the Transformer encoder,
and $\mathbf{h}_{dec}^t$ is the output of the Transformer decoder at timestep $t$.
$p_{v}(w)$ is the standard softmax probability for word $w$ over the vocabulary
($p_{v}(w) = 0$ if $w$ is an OOV word).
$\boldsymbol{\alpha}$ is the copy attention over the input sequence,
\emph{AttPooling} is the attentive pooling of encoder outputs $\mathbf{h}_{enc}$ with distribution $\boldsymbol{\alpha}$,
and $p_{gen}$ denotes the probability of generating from the fixed vocabulary.
$p(w)$ is a linear interpolation of the generation and copying probabilities.

\end{document}